# Periocular Biometrics: A Modality for Unconstrained Scenarios

**Fernando Alonso-Fernandez**
School of Information Technology, Halmstad University, Sweden

**Josef Bigun**
School of Information Technology, Halmstad University, Sweden

**Julian Fierrez**
School of Engineering, Universidad Autonoma de Madrid, Spain

**Naser Damer**
Fraunhofer Institute for Computer Graphics Research IGD, Darmstadt, Germany
Department of Computer Science, TU Darmstadt, Darmstadt, Germany

**Hugo Proença**
Instituto de Telecomunicações, University of Beira Interior, Portugal

**Arun Ross**
Department of Computer Science and Engineering, Michigan State University, USA

*Abstract*—Periocular refers to the externally visible region of the face that surrounds the eye socket. This feature-rich area can provide accurate identification in unconstrained or uncooperative scenarios, where the iris or face modalities may not offer sufficient biometric cues due to factors such as partial occlusion or high subject-to-camera distance. The COVID-19 pandemic has further highlighted its importance, as the ocular region remained the only visible facial area even in controlled settings due to the widespread use of masks. This paper discusses the state of the art in periocular biometrics, presenting an overall framework encompassing its most significant research aspects, which include: (a) ocular definition, acquisition, and detection; (b) identity recognition, including combination with other modalities and use of various spectra; and (c) ocular soft-biometric analysis. Finally, we conclude by addressing current challenges and proposing future directions.

■ **THE OCULAR REGION** consists of several organs, including the cornea, pupil, iris, sclera, lens, retina, and eyelids, among others (Figure 1). Among these, the iris, sclera, retina, and periocular entities have been studied as biometric modalities, particularly the iris [1]. However, iris recognition systems primarily operate with near-infrared (NIR) illumination and controlled close-up acquisition. In visible (VIS) illumination, performance significantly degrades. Moreover, real-world conditions present challenges such as occlusion, subjects' pose, unfavorable illumination, and low resolution, which may even hinder iris localization or the acquisition of suitable iris images. Face recognition technologies have also seen significant progress over the last two decades, but unconstrained recognition remains difficult. Partial faces have become an issue even in controlled setups during the pandemic due to the mandatory use of masks in some places, negatively impacting state-of-the-art facial recognition systems [2].

In this context, periocular biometrics has rapidly evolved as a promising approach for unconstrained biometrics. Several recent survey papers [1], [3], [4], including those specifically addressing mask-related challenges [5], have contributed to this field. Several competitions have also been organized over the years [6]. The ocular region by itself has demonstrated effectiveness in identity recognition [3], soft-biometrics estimation [7], and expression analysis [5]. It appears both in iris and face images, so it is easily obtainable with existing sensors. It also remains visible at various distances, even when face occlusion occurs due to close acquisition (e.g., selfie [8]) or when the stand-off distance prevents high-resolution iris imaging. Moreover, in many un-cooperative scenarios, it may be the only visible area, (involuntarily or voluntarily (e.g., criminals concealing their faces). Even in cooperative situations, the use of facial masks during the pandemic obstructs most of the face, revealing only the eyes and their immediate surrounding, affecting all kinds of applications employing face technologies in security, healthcare, border control, education, and other domains.

Our scope is thus the ocular region. This paper aims to provide insights into key aspects of periocular biometrics, covering the entire pipeline from the definition of the ocular region to its acquisition, detection, and identity recognition. We also discuss aspects like combining with other modalities to enhance recognition accuracy (typically face or iris), recognition in different spectra [9], or estimating demographic attributes (gender, age, and ethnicity) from ocular images. The paper concludes by highlighting current challenges and future directions. Existing recent surveys [1], [4], [5], [6] primarily describe specific feature methods, algorithms, datasets, and benchmarks. This paper takes a more practical approach, focusing on technical aspects but omitting detailed algorithmic specifics (found in referenced surveys). We only mention the achieved accuracy for a specific task whenever relevant, referring interested readers to papers dedicated to systematic reviews of datasets and benchmarks [6]. In addition, due to the journal's limitation in the number of references, original works cannot always be cited directly. In such cases, we refer readers to survey papers for comprehensive details about the mentioned issues. This mostly applies to older papers.

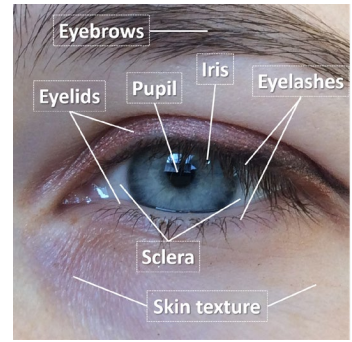

**Figure 1.** Eye image labeled with some parts of the ocular region.

## THE PERIOCULAR REGION: DEFINITION, ACQUISITION, AND DETECTION

The medical definition of "periocular", according to the Merriam-Webster dictionary, is "surrounding the eyeball but within the orbit". In biometrics, the term is used loosely to refer to the externally visible region of the face around the eye socket, and sometimes is used interchangeably with the term "ocular". Thus, periocular systems employ the entire eye image as input, as depicted in Figure 1. While components like the iris and sclera are present, they are not necessarily used in isolation or may not have sufficient quality to be processed reliably on their own. It is important to note that there is no standardized definition for the periocular region of interest (ROI), resulting in variations across papers. Additionally, some authors use the eye center as reference, while others use the eye corners, which are less sensitive to gaze variations [4].

Initial research employed face or iris datasets due to limited availability of periocular ones. Sensing devices included digital cameras, webcams, video cameras, or close-up iris sensors. As research progressed, specific datasets emerged.

A detailed description of face, iris, and periocular databases can be found in existing surveys [1], [3], [5], [6] and newer papers [10]. Figure 2 shows sample images from periocular databases and the best-reported accuracy on those datasets. They are categorized into NIR and VIS databases. Most VIS databases (CSIP, MICHE, VSSIRIS, VISOB, UFPR) have been captured using mobile devices by volunteers themselves, introducing variabilities like blur, defocus, reflections, eyeglasses, off-angle gaze, pose, makeup, or expression. These databases also include different sensors and environments (indoor/outdoor, natural light/office light, etc.) Some are with long-range devices (FOCS, CASIA distance) or zoomable digital cameras (UBIPr), and there are a few multiple spectra sets, enabling cross-spectral periocular analysis [9]. Also, although several sets involve different acquisition distances (e.g. MIR 2016, CASIA Iris Mobile, UBIPr), subjects usually stand at predetermined stand-off distances. The only database with true mobility is FOCS, with subjects walking through an acquisition portal. This introduces significant challenges, such as motion blur or scale changes, resulting in lower accuracy (EER: 18.8%) compared to other databases. Such result highlights on-the-move operation as an open challenge in periocular recognition. Certain databases serve specific purposes. For example, CMPD contains subjects before/after cataract surgery, a common disease among elderly people. The reported results comparing images before/after surgery (RR: 30.10%) indicate that cataract surgery significantly impacts periocular recognition.

In research studies, automatic detection of the ocular region has not been a primary focus. Instead, the emphasis has been on feature extraction for recognition or other tasks such as soft-biometrics [11]. Initially, manual marking of the region of interest or extraction after full-face detection was commonly used. In comparison to face detection research, which has spanned several decades, very few methods have been proposed to locate the eyes directly without the support of the nose-to-chin region [3]. State-of-the-art face detectors, including those based on *deep-learning* (DL), are designed to detect the entire face. Occlusion is present in training databases but is not specifically controlled, nor are the methods trained or evaluated on their capabilities when only the ocular area is visible. Occluded face detection has been attracting research recently, including methods that locate the visible parts of the face [12]. However, these approaches primarily focus on analyzing face subregions (mouth, nose, etc.) to infer the potential location of the full face. Detecting the ocular region directly without relying on full-face detection or a systematic analysis of expected subparts is thus an under-researched area.

## PERIOCULAR BIOMETRICS AS A STANDALONE MODALITY

One of the earliest papers on periocular biometrics was by Park et al. in 2009 [13]. Simple texture operators were used to encode the periocular region. Subsequently, in 2011 (see [3]), a more detailed analysis was conducted, exploring the effectiveness of incorporating eyebrows, the possibility of fusing face and periocular modalities, the impact of varying pose and illumination, masking the iris and eye region, etc. In particular, the authors demonstrated the benefits of the periocular modality when the face was partially occluded.

Since then, various methods have been employed to encode the periocular region. These include classical texture operators (LBP, BSIF, BRISK, HOG, SIFT, SURF, etc.) and filters (Gabor, Leung-Malik, etc.) [3], [4]. The importance of different elements within the ocular region and the size of the region around the eye have been subjects of scrutiny as well [5]. For example, texture and color information (skin, wrinkles, pores, etc.) are more useful in the visible spectrum. In the near-infrared spectrum, such cues are obfuscated (see Figure 2), so ocular geometry information (eyelids, lashes, brows, etc.) becomes more relevant.

More recently, due to the prevalence of deep learning (DL) techniques, Convolutional Neural Networks (CNNs) have gained popularity [5], either employing off-the-shelf CNN features or training networks using autoencoders or attention mechanisms to guide the network to focus on relevant regions like eyebrows and eyelashes [10]. The current state of the art is given by DL models. However, one drawback of these models is their reliance on large-scale databases, which are currently lacking in periocular research. Most of the datasets mentioned in the previous section contain only a few thousand images at most. The largest available dataset (VISOB v1.0) comprises 158k images, which is one or two orders of magnitude lower compared to the datasets available in other modalities, such as face biometrics. Therefore, the scarcity of large-scale periocular databases poses a challenge in advancing the field of periocular biometrics.

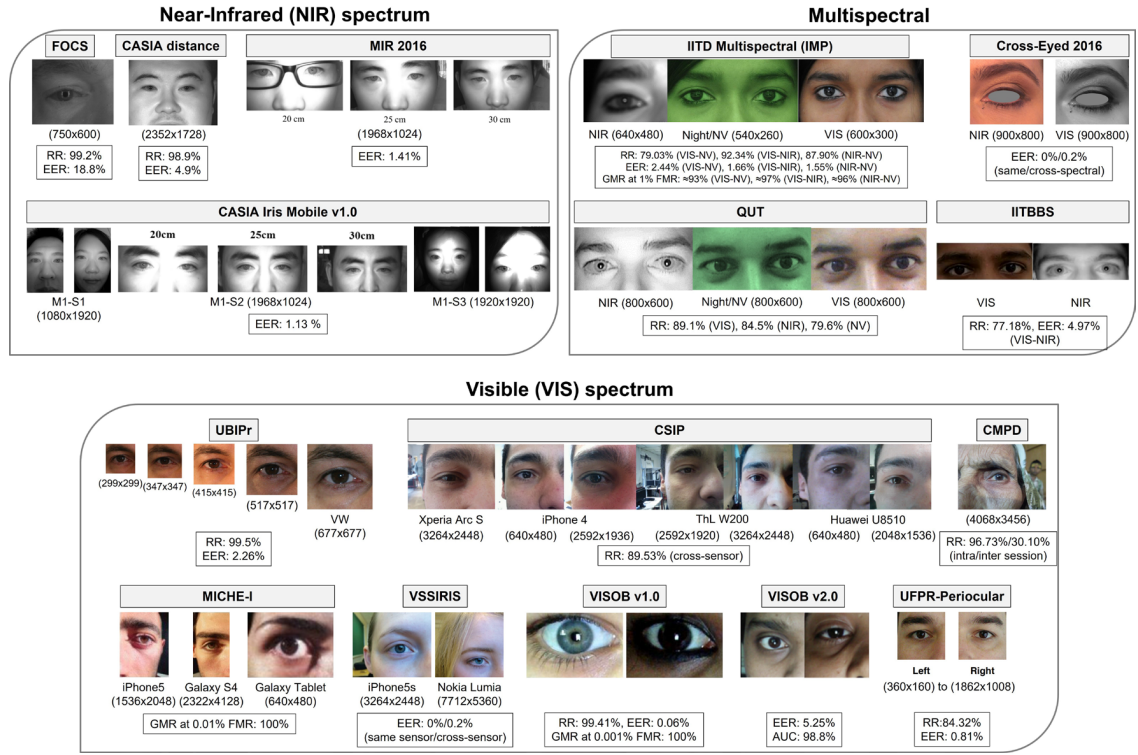

**Figure 2.** Samples of periocular databases and the best results in the literature with the periocular modality (results adapted from references in [5], [6] and other newer references [10]). RR: Recognition Rate. EER: Equal Error Rate. GMR: Genuine Matching Rate. FMR: False Matching Rate. AUC: Area Under the Curve.

## COMBINATION WITH OTHER MODALITIES

From the beginning, the periocular region has been considered valuable for unconstrained data acquisition in visual surveillance scenarios [14]. However, data obtained in such settings often lack intra-subject permanence and discriminability between subjects, which is the main rationale for fusing the periocular region with other biometric traits to improve the overall performance.

The iris, due to its biological proximity, is frequently chosen for fusion. This combination is especially useful when the iris has insufficient quality due to reflections, off-axis gaze, motion, low resolution, etc. Different texture descriptors are used in existing works, such as classical Gabor kernels for the iris, and LBP, HOG, or Leung-Malik for the periocular region [5]. Fusion is typically performed at the score level. More recently, deep learning models have also been employed, leveraging joint attention mechanisms to learn relevant features of each region.

Fusing descriptions from the entire face and the periocular region is also common. This is beneficial when the face is partially occluded, exhibits significant pose variation, or is captured at a close distance. As in the case of the iris, the idea is to obtain independent feature representations from the face and periocular region, delimited using hard-attention mechanisms and fused at the feature or score levels. Earlier attempts included traditional features like Gabor wavelets, LBP, HOG, or SIFT [3]. Recent works explore DL solutions such as shared backbones or siamese models with an independent stream for each one.

Lastly, the sclera region should also be mentioned as another trait frequently advocated for fusion with the periocular region [5]. Various features and methods for sclera detection and segmentation have been proposed over the years. The sclera is particularly advantageous in the VIS spectrum, where its prominent blood vessels are easily visible.

In conclusion, most studies highlight the benefits of fusing periocular information with other neighboring traits. The exception is due to Proenç̧a and Neves [15], who argued that recognition performance in the VIS spectrum is optimized when components within the ocular globe (iris and sclera) are discarded, and the recognizer's response is solely based on the surrounding eye information.

## RECOGNITION IN DIFFERENT SPECTRA

Image-based biometrics utilize camera sensors that measure different light wavelength ranges. The three main considered spectra are visible (VIS), near-infrared (NIR), and infrared (IR). Each poses advantages and restrictions for periocular biometric systems and application scenarios [16]. For example, VIS enables the use of many existing built-in cameras, offers high detail, and is suitable for scenarios such as self-verification and surveillance. NIR illumination, on the other hand, reveal details unseen in the VIS spectrum (e.g. in iris recognition, as the effect of melanin is negligible under NIR), is less sensitive to illumination variations, and is comfortable to the human eye because is not perceivable. Such properties make NIR suitable for periocular recognition in combination with iris or under illumination-sensitive scenarios such as head-mounted displays. However, it requires an active NIR invisible illumination source. IR imaging, also known as thermal imaging, provides lower information details and is more sensitive to environmental variations, making it less suitable for periocular recognition.

At the algorithmic level, these spectra capture different sets of information from the periocular region. The two main periocular recognition challenges in this scope are (1) accurate recognition under each spectrum to adapt to different use-cases and (2) accurate recognition in a cross-spectral setting where the reference and probe are captured under different spectra [9].

Recognition in the VIS spectrum is motivated by using existing general-purpose capture devices for self-verification (e.g. smartphones [8]) or surveillance scenarios, including occluded or masked faces. Numerous databases have been collected to develop VIS periocular recognition (Figure 2). NIR recognition, on the other hand, is driven by capture devices used for iris recognition and scenarios where VIS is not applicable, such as head-mounted displays in augmented and virtual reality applications [17]. Solutions for intra-spectrum periocular recognition (NIR or VIS) are technically similar, utilizing handcrafted features, deeply learned representations, and their fusion [5]. This interest in advancing intra-spectral periocular biometrics led to the organization of a series of competitions, including the VISOB 1.0 and VISOB 2.0 events [6].

Many applications restrict the biometric reference to be captured under one spectrum, but require the ability to match probes captured under other spectra. This raises the challenge of cross-spectral periocular recognition. Two main directions have been followed in this regard: direct comparison using features less sensitive to spectral changes or specifically learned features that produce similar representations for NIR and VIS images of the same identity [10], and generative transformation of the probe into the reference domain, where an intra-spectral recognition algorithm is applied [18]. Given its highly challenging nature, competitions like the Cross-Eyed series have focused on attracting novel solutions for cross-spectral periocular recognition [6], [9]. As it can be observed in Figure 2, accuracy in cross-spectral datasets (top left) is typically lower than in intra-spectral NIR or VIS operation.

## DEMOGRAPHICS FROM OCULAR IMAGES

Soft-biometrics refer to *ancillary* information such as age, gender, race, handedness, height, weight, hair color, etc., that can help when recognizing a person [11]. Among these, demographic indicators (gender, age, ethnicity) have special relevance because they can be linked to unde-sired discrimination between population groups [19]. Soft-biometrics can be computed from the body silhouette or the face, although some have suggested computing them from fingerprints, iris, or handwriting.

In controlled scenarios, face or iris biometrics can be very effective to recognize an individual. But under difficult covariates in real-world conditions (occlusion, subjects' pose, illumination, resolution, etc.), demographic attributes can be retrieved with a higher probability of success. They can be used in isolation, or complement the inconclusive decision of stronger biometric modalities. For example, combining soft biometrics with periocular features has shown enhanced overall recognition performance [20].

Demographic attributes also find applications in targeted advertising, searching for individuals based on specific attributes, age-related access control, or child pornography detection. Although demographic estimation is often seen as relatively easy, extracting such attributes *in-the-wild* is challenging. However, research in this area primarily focuses on good-quality data and frequently uses the entire face, despite likely occlusions in unconstrained setups such as forensics or surveillance [21].

Gender estimation (male/female) is the most widely studied attribute and considered the easiest due to its binary nature. Initial works can be traced back to 2010 [3], cropping the ocular area from well-established face recognition databases. Later works incorporated selfie images from smartphones and leveraged learned features via CNNs. Recent works achieve accuracies above 80-90% in gender estimation [7].

Ethnicity estimation poses challenges in defining classes consistently across databases, and some may be severely under-represented. Most databases have only two or three ethnic classes since they were not specifically acquired for ethnicity estimation. Initial works can be also traced back to 2010, but the literature on ocular ethnicity is much less compared to gender. Accuracies above 80-90% are common as well, but comparing results between works is difficult due to differences in classes across databases.

Age is considered the most complex attribute due to internal (genetics) and external (health, stress, lifestyle...) factors influencing the aging process. Comparatively, it is the most under-researched demographics with ocular data. Classes are often discretized (e.g. children, teens, adults...), achieving higher performance compared to estimating the exact age, and allowing customization to requirements (e.g. minors/non-minors). Pioneering works in 2015 used controlled data, followed later by selfies and in-the-wild imagery. Recent works barely exceed a 60% [7] accuracy, highlighting the difficulty of the task. It is also common to report the *1-off* accuracy, which considers classifications for adjacent age groups as correct. This more tolerant framework provides accuracies above 80%.

## CONCLUSION, CHALLENGES, AND FUTURE DIRECTIONS

In the last decade, the periocular modality has rapidly evolved, surpassing face in case of occlusion or iris under low resolution. Periocular is the region around the eye, comprising the sclera, eyelids, lashes, brows, and surrounding skin. With a surprisingly high discrimination ability, it requires less constrained acquisition than the iris texture. It remains visible at various distances, even with partial face occlusion due to close distance, or low resolution due to long distances. This makes it suitable for unconstrained or uncooperative scenarios where iris or face recognition may struggle. The periocular modality gained renewed attention during the pandemic as masks left the ocular region as the only visible facial area, even in controlled situations. Apart from personal recognition, periocular biometrics have been used for demographics [7] or expression estimation [5]. Figure 3 provides a graphical summary of periocular biometrics, including aspects mentioned during the paper and challenges discussed in the present section.

Despite the advances mentioned in this paper, several research challenges remain. Questions about the optimal size of the periocular ROI and the minimum resolution required for recognition are still open [4], [5]. The lack of a standardized definition for the periocular region leads to variations in the employed ROI across studies. For example, some studies exclude the sclera, iris, and pupil. Additionally, some consider the two eyes as a single instance, while others treat each eye as a separate unit. Large-scale datasets and benchmarks are needed as well to leverage data-hungry deep-learning schemes and promote further research and replication [1], [6]. A recent concern affecting all biometric modalities is demographic bias and fairness, where certain demographic groups may experience lower classification accuracy. This is not exclusive to biometrics, but it is common in automated decision-making systems [19]. Although face algorithms have attracted the majority of public attention in this regard, proper mitigation measures are also needed in ocular biometrics. The increasing use of deep-learning solutions also raises questions about explainability, due to their black-box nature, i.e. why a recognition system makes certain decisions.

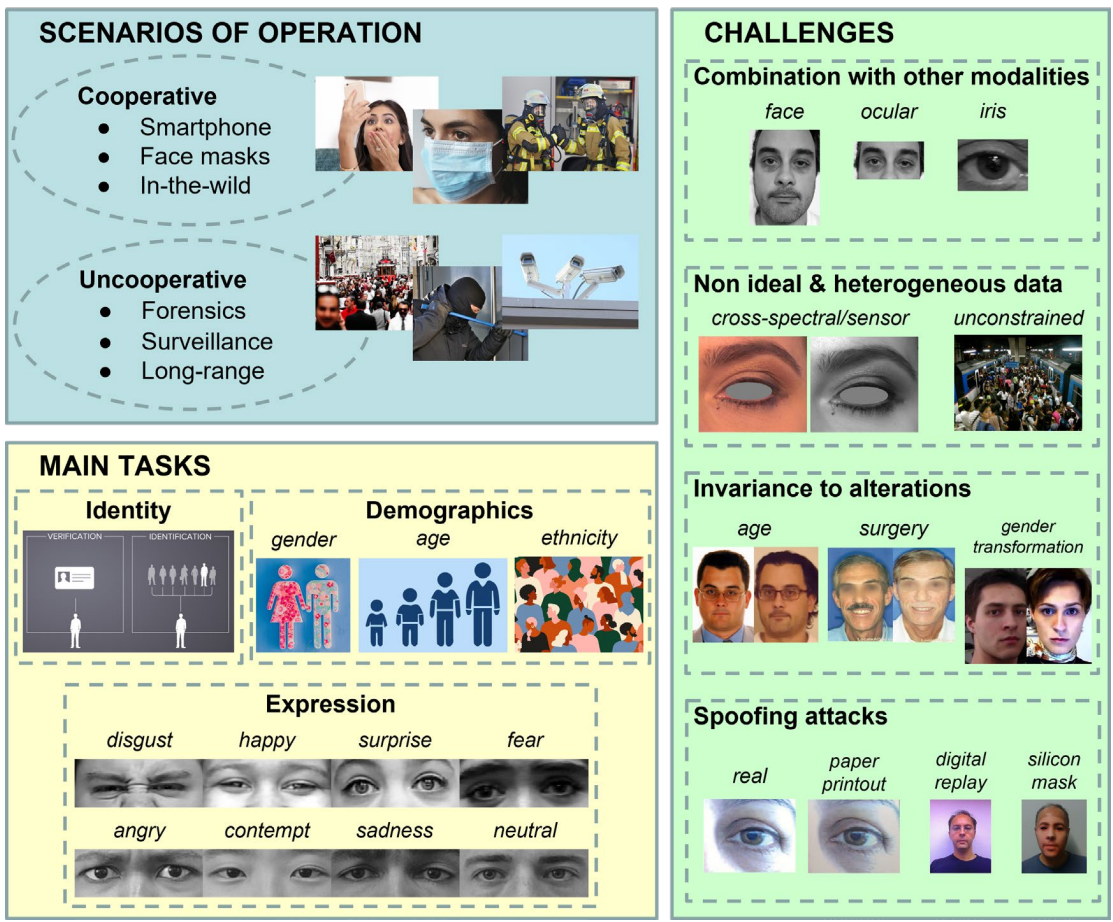

**Figure 3.** Graphical summary of different aspects of significance in periocular biometrics. Top left: potential scenarios of operation. Top bottom: main tasks where periocular images can be useful. Right: Some other challenges affecting periocular biometrics.

Other challenges that are worth mentioning include the following:

- **Acquisition of high-quality images**. This is vital to any biometric modality. Most periocular datasets come from mobile devices or cooperative subjects zoomed from a close distance [6]. Factors like less cooperative scans, motion, and larger stand-off distances are under-researched. Several hardware solutions have been proposed, such as hyper-focal or light-field sensors that fuse images with different focal lengths [1], or near-infrared walking portals that capture approaching individuals [3]. However, they come with extra cost or increased sensor size, making them impractical for consumer or forensic applications.

- **Smartphone authentication**. The pandemic

accelerated digital service provision through personal devices, which have become data hubs containing sensitive information. Their inherent *on-the-move* conditions cause imaging difficulties that can severely degrade performance. Also, device usage in diverse environments introduces variability in pose, illumination, background, etc. The availability of different device models with unknown camera specifications further complicates operations. Operation under such circumstances is referred to as cross-device (different devices) or cross-environment (different acquisition environments), which demands mitigation methods to minimize adverse effects on performance [1].

- **Heterogeneous operation**. Despite impressive periocular recognition performance under ideal conditions, maintaining it across different sensors, spectral ranges, and resolutions remains challenging [5], [6]. Part of this challenge is related to the lack of large-scale multi-spectral databases suitable to train deep neural networks with millions of parameters. The largest ocular database (VISOB, with 550 subjects/158k images from 3 sensors) contrasts with the millions of images available to train, for example, face recognition models. This motivates recent efforts [22] that explore identity-aware synthetic periocular data as a replacement for authentic data. Generative methods have shown impressive ability in creating realistic synthetic data across various applications. In biometrics, they address privacy concerns tied to obtaining and publicly sharing benchmark databases while providing sufficient data for training deep-learning methods.

- **Deployability**. Recent works show the superior accuracy and generalizability of deep learning-based periocular recognition compared to handcrafted features. However, such models impose high requirements in model size and computational complexity that makes them undeployable on resource-critical consumer devices. This motivates future research to work towards harvesting the knowledge learned with deeper (larger) models and transferring it into more deployable models with reduced size and computational complexity, while maintaining performance [22].

- **Invariance to age and other alterations**. Being a relatively recent addition to the family of biometric traits, various factors can influence periocular recognition performance, such as facial expressions, potential forgery through surgical procedures, and - in particular - long-term stability of periocular features, i.e., invariance to aging. Although the periocular region is relatively more stable and less affected than the entire face, few studies have examined their impact on periocular methods [3], [5]. Analyzing these factors is crucial to enhance confidence in periocular-based recognition systems and establish it as a viable biometric recognition solution.

- **Spoofing attacks**. In parallel with the popularity of biometrics systems, their security against attacks has become paramount. The most common attack, presentation attack (also known as spoofing), consists in presenting a fake biometric sample to the sensor. This has received extensive attention with face and iris modalities to detect e.g. silicon masks, printouts, contact lenses, or digital replays. Although several works exist with ocular images [23], the amount is much more limited [5], [6].

## ACKNOWLEDGMENT

The authors thank support from: the Swedish Research Council VR and Innovation Agency VINNOVA (F. Alonso-Fernandez, J. Bigun); the projects BBforTAI (PID2021-127641OB-I00 MICINN/FEDER) and HumanCAIC (TED2021-131787B-I00 MICINN) (J. Fierrez); the German Federal Ministry of Education and Research and the Hessian Ministry of Higher Education, Research, Science and the Arts within their joint support of the National Research Center for Applied Cybersecurity ATHENE (N. Damer); FCT/MCTES through national funds and co-funded by EU funds under the project UIDB/50008/2020 (H. Proenc¸a); the US National Science Foundation, Award 1841517 (CITeR) (A. Ross). The data handling in Sweden was enabled by the National Academic Infrastructure for Supercomputing in Sweden (NAISS).

# REFERENCES

1. A. Rattani and R. Derakhshani, "Ocular biometrics in the visible spectrum: A survey," *Image and Vision Computing*, vol. 59, pp. 1 – 16, 2017.
2. M. Ngan, P. Grother, and K. Hanaoka, "Part 6b: Face recognition accuracy with face masks using post-COVID-19 algorithms," *NISTIR 8331 - https://pages.nist.gov/frvt/html/frvt_facemask.html - last updated: March 4, 2021*, 2020.
3. F. Alonso-Fernandez and J. Bigun, "A survey on periocular biometrics research," *Pattern Recognition Letters*, vol. 82, pp. 92–105, 2016.
4. P. Kumari and K. Seeja, "Periocular biometrics: A survey," *Journal of King Saud University - Computer and Information Sciences*, vol. 34, no. 4, pp. 1086–1097, 2022.
5. R. Sharma and A. Ross, "Periocular biometrics and its relevance to partially masked faces: A survey," *Computer Vision and Image Understanding*, vol. 226, p. 103583, 2023.
6. L. A. Zanlorensi, R. Laroca, E. Luz, A. S. Britto, L. S. Oliveira, and D. Menotti, "Ocular recognition databases and competitions: a survey," *Artificial Intelligence Review*, vol. 55, no. 1, pp. 129–180, 2022.
7. F. Alonso-Fernandez, K. Hernandez-Diaz, S. Ramis, F. J. Perales, and J. Bigun, "Facial masks and soft-biometrics: Leveraging face recognition CNNs for age and gender prediction on mobile ocular images," *IET Biometrics*, vol. 10, no. 5, pp. 562–580, 2021.
8. F. Alonso-Fernandez, R. A. Farrugia, J. Fierrez, and J. Bigun, "Super-resolution for selfie biometrics: Introduction and application to face and iris," in *Selfie Biometrics*, A. Rattani, R. Derakhshani, and A. Ross, Eds. Springer, 2019, pp. 105–128.
9. F. Alonso-Fernandez, K. Raja, R. Raghavendra, C. Busch, J. Bigun, R. Vera-Rodriguez, and J. Fierrez, "Cross-sensor periocular biometrics for partial face recognition in a global pandemic: Comparative benchmark and novel multialgorithmic approach," *Information Fusion*, vol. 83-84, pp. 110–130, July 2022.
10. S. S. Behera, N. B. Puhan, and S. S. Mishra, "Perturbed attention-assisted siamese network for cross-spectral periocular recognition," *IEEE Transactions on Biometrics, Behavior, and Identity Science*, vol. 4, no. 2, pp. 210–221, 2022.
11. E. Gonzalez-Sosa, J. Fierrez, R. Vera-Rodriguez, and F. Alonso-Fernandez, "Facial soft biometrics for recognition in the wild: Recent works, annotation and cots evaluation," *IEEE Trans. on Information Forensics and Security*, vol. 13, no. 8, pp. 2001–2014, August 2018.
12. D. Zeng, R. Veldhuis, and L. Spreeuwers, "A survey of face recognition techniques under occlusion," *IET Biometrics*, vol. 10, no. 6, pp. 581–606, 2021.
13. U. Park, A. Ross, and A. K. Jain, "Periocular biometrics in the visible spectrum: A feasibility study," in *IEEE 3rd International Conference on Biometrics: Theory, Applications, and Systems*, 2009.
14. H. Proença and J. C. Neves, "Visible-wavelength iris/periocular imaging and recognition surveillance environments," *Image and Vision Computing*, vol. 55, pp. 22–25, 2016.
15. ——, "Deep-PRWIS: Periocular recognition without the iris and sclera using deep learning frameworks," *IEEE Transactions on Information Forensics and Security*, vol. 13, no. 4, pp. 888–896, 2018.
16. M. Moreno-Moreno, J. Fierrez, and J. Ortega-Garcia, "Biometrics beyond the visible spectrum: Imaging technologies and applications," in *Proceedings of BioID-Multicomm*, ser. LNCS, J. Fierrez, J. Ortega-Garcia, A. Esposito, A. Drygajlo, and M. Faundez-Zanuy, Eds., vol. 5707. Springer, September 2009, pp. 154–161.
17. F. Boutros, N. Damer, K. B. Raja, R. Ramachandra, F. Kirchbuchner, and A. Kuijper, "Iris and periocular biometrics for head mounted displays: Segmentation, recognition, and synthetic data generation," *Image Vis. Comput.*, vol. 104, p. 104007, 2020.
18. K. B. Raja, N. Damer, R. Ramachandra, F. Boutros, and C. Busch, "Cross-spectral periocular recognition by cascaded spectral image transformation," in *IEEE International Conference on Imaging Systems and Techniques, IST*, 2019.
19. I. Serna, A. Morales, J. Fierrez, and N. Obradovich, "Sensitive loss: Improving accuracy and fairness of face representations with discrimination-aware deep learning," *Artificial Intelligence*, vol. 305, p. 103682, April 2022.
20. V. Talreja, N. M. Nasrabadi, and M. C. Valenti, "Attribute-based deep periocular recognition: Leveraging soft biometrics to improve periocular recognition," in *2022 IEEE/CVF Winter Conference on Applications of Computer Vision (WACV)*, 2022, pp. 1141–1150.
21. F. Becerra-Riera, A. Morales-González, and H. Méndez-Vázquez, "A survey on facial soft biometrics for video surveillance and forensic applications," *Artif. Intell. Rev.*, vol. 52, no. 2, p. 1155–1187, 2019.
22. J. N. Kolf, J. Elliesen, F. Boutros, H. Proença, and N. Damer, "SyPer: Synthetic periocular data for quantized light-weight recognition in the NIR and visible

**Fernando Alonso-Fernandez** is a docent and an Associate Professor at Halmstad University, Sweden. He received the M.S./Ph.D. degrees in telecommunications from Universidad Politecnica de Madrid, Spain, in 2003/2008. His research interests include AI for biometrics and security, signal and image processing, feature extraction, pattern recognition, and computer vision. He is an IEEE member. Contact him at feralo@hh.se.

**Josef Bigun** is a Full Professor of the Signal Analysis Chair at Halmstad University, Sweden. He received the M.S./Ph.D. degrees from Linköping University, Sweden, in 1983/1988. His scientific interests broadly include computer vision, texture and motion analysis, biometrics, and the understanding of biological recognition mechanisms. He is an IEEE fellow. Contact him at josef.bigun@hh.se.

**Julian Fierrez** is a Professor at Universidad Autónoma de Madrid, Spain. He received the M.Sc./Ph.D. degrees in telecommunications engineering from the Universidad Politécnica de Madrid, Spain, in 2001/2006. His research interests include signal and image processing, HCI, responsible AI, and biometrics for security and human behavior analysis. He is an IEEE member. Contact him at julian.fierrez@uam.es.

**Naser Damer** is a Senior Researcher with Fraunhofer IGD, performing research management, applied research, scientific consulting, and system evaluation. He received the Ph.D. degree in computer science from TU Darmstadt in 2018. His main research interests lie in biometrics, machine learning, and information fusion. He is an IEEE member. Contact him at Naser.Damer@igd.fraunhofer.de.

**Hugo Proença** is a Full Professor at the Department of Computer Science, University of Beira Interior, Portugal. He received the B.Sc., M.Sc., and Ph.D. degrees in 2001, 2004, and 2007, respectively. His research interest encompasses biometrics and visual surveillance. He is an IEEE senior member. Contact him at hugomcp@di.ubi.pt.

**Arun Ross** is currently the Martin J. Vanderploeg Endowed Professor in the College of Engineering and a Professor with the Department of Computer Science and Engineering, at Michigan State University. He received the B.E. (Hons.) degree in computer science from BITS Pilani, India, and the M.S. and Ph.D. degrees in computer science and engineering from Michigan State University. His research interests include AI, biometrics, computer vision, machine learning, and pattern recognition. He is an IEEE senior member. Contact him at rossarun@cse.msu.edu.